\documentclass[11pt,a4paper]{article}

\usepackage[utf8]{inputenc} 
\usepackage{textcomp}

\usepackage[hyperref]{emnlp2018}
\usepackage{todos}

\usepackage{times}
\usepackage{latexsym}

\usepackage{amsmath}
\usepackage{amssymb}

\usepackage{bm}
\usepackage{color}
\usepackage{multirow}
\usepackage{subfig}
\usepackage{graphicx, wrapfig}
\usepackage{arydshln}
\usepackage{balance}

\DeclareMathOperator*{\argmax}{arg\,max}

\aclfinalcopy 

\DeclareSymbolFont{extraup}{U}{zavm}{m}{n}
\DeclareMathSymbol{\varheart}{\mathalpha}{extraup}{86}
\DeclareMathSymbol{\vardiamond}{\mathalpha}{extraup}{87}

\begin{document}

\title{Multi-Head Attention with Disagreement Regularization}

\author{Jian Li$^{1,2}$~~~~~~Zhaopeng Tu$^3$\thanks{~~~Zhaopeng Tu is the corresponding author of the paper. This work was mainly conducted when Jian Li and Baosong Yang were interning at Tencent AI Lab.}~~~~~~Baosong Yang$^4$~~~~~~Michael R. Lyu$^{1,2}$~~~~~~Tong Zhang$^3$\\
  {\normalsize $^1$Department of Computer Science and Engineering, The Chinese University of Hong Kong}\\
  {\normalsize $^2$Shenzhen Research Institute, The Chinese University of Hong Kong}\\
  {\normalsize \tt $^{1,2}$\{jianli,lyu\}@cse.cuhk.edu.hk}\\
  {\normalsize $^3$Tencent AI Lab}    ~~~~~~~~~~~~~~~~~~~~~~~~~~~~~~~~~~~~~~~~~~~~~~~~~  {\normalsize $^4$University of Macau}\\
  {\normalsize \tt $^3$\{zptu,bradymzhang\}@tencent.com}    ~~~~~~~~~~~~~~~~~~  {\normalsize \tt $^4$nlp2ct.baosong@gmail.com}}

\maketitle
\begin{abstract}

Multi-head attention is appealing for the ability to jointly attend to information from different representation subspaces at different positions. In this work, we introduce a disagreement regularization to explicitly encourage the diversity among multiple attention heads. 
Specifically, we propose three types of disagreement regularization, which respectively encourage the subspace, the attended positions, and the output representation associated with each attention head to be different from other heads.
Experimental results on widely-used WMT14 English$\Rightarrow$German and WMT17 Chinese$\Rightarrow$English translation tasks demonstrate the effectiveness and universality of the proposed approach. 


\end{abstract}

\section{Introduction}

Attention model is now a standard component of the deep learning networks, contributing to impressive results in neural machine translation~\cite{Bahdanau:2015:ICLR,Luong:2015:EMNLP}, image captioning~\cite{xu2015show}, speech recognition~\cite{chorowski2015attention}, among many other applications. Recently,~\newcite{Vaswani:2017:NIPS} introduced a multi-head attention mechanism to capture different context with multiple individual attention functions. 

One strong point of multi-head attention is the ability to jointly attend to information from {\em different} representation subspaces at {\em different} positions.
However, there is no mechanism to guarantee that different attention heads indeed capture distinct features.
In response to this problem, we introduce a disagreement regularization term to explicitly encourage the diversity among multiple attention heads.
The disagreement regularization serves as an auxiliary objective to guide the training of the related attention component.

Specifically, we propose three types of disagreement regularization, which are applied to the three key components that refer to the calculation of feature vector using multi-head attention.
Two regularization terms are respectively to maximize cosine distances of the input subspaces and output representations, while the last one is to disperse the positions attended by multiple heads with element-wise multiplication of the corresponding attention matrices.
The three regularization terms can be either used individually or in combination.


We validate our approach on top of advanced \textsc{Transformer} model~\cite{Vaswani:2017:NIPS} for both English$\Rightarrow$German and Chinese$\Rightarrow$English translation tasks.
Experimental results show that our approach consistently improves translation performance across language pairs. 
One encouraging finding is that \textsc{Transformer-Base} with disagreement regularization achieves comparable performance with \textsc{Transformer-Big}, while the training speed is nearly twice faster.

\section{Background: Multi-Head Attention}

\begin{figure}[h]
  \centering
  \includegraphics[width=0.4\textwidth]{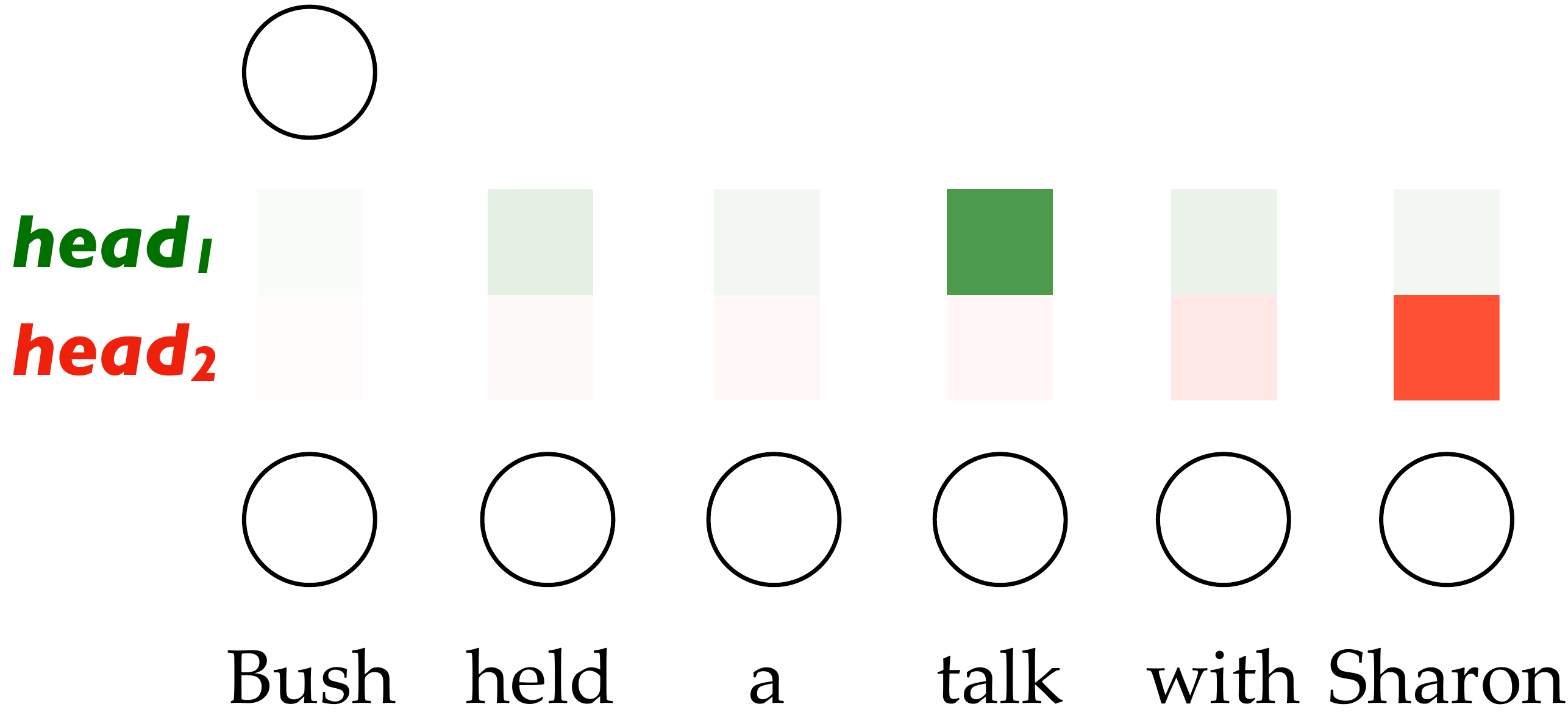}
\caption{Illustration of the multi-head attention, which jointly attends to different representation subspaces (colored boxes) at different positions (darker color denotes higher attention probability).}
\label{fig:multi-head}
\end{figure}


Attention mechanism aims at modeling the strength of relevance between representation pairs, such that a representation is allowed to build a direct relation with another representation.
Instead of performing a single attention function,~\newcite{Vaswani:2017:NIPS} found it is beneficial to capture different context with multiple individual attention functions.  Figure~\ref{fig:multi-head} shows an example of a two-head attention model. For the query word ``Bush'', green and red head pay attention to different positions of ``talk'' and ``Sharon'' respectively. 

Attention function softly maps a sequence of query $Q=\{Q_1,\dots,Q_N\}$ and a set of key-value pairs $\{K,V\}=\{(K_1,V_1),\dots,(K_M,V_M)\}$ to outputs.
More specifically,  multi-head attention model first transforms $Q$, $K$, and $V$ into $H$ subspaces, with different, learnable linear projections,  namely:
\begin{equation*}
  Q^h, K^h, V^h   = QW_h^{Q}, KW_h^{K}, VW_h^{V},  
\end{equation*}
where $\{Q^h, K^h, V^h\}$ are respective the query, key, and value representations of the $h$-th head. $\{W_h^{Q}, W_h^{K}, W_h^{V}\} \in \mathbb{R}^{d \times d_k}$ 
denote parameter matrices, $d$ and $d_k$ 
represent the dimensionality of the model and 
its subspace. Furthermore, $H$ attention functions are applied in parallel to produce the output states $\{O^1,\dots,O^H\}$, among them:
\begin{equation*}
  O^h  = A^h V^h \ \  with \ \  A^h = softmax(\frac{Q^h {K^h}^T}{\sqrt{d_k}}).  
\end{equation*}
 Here $A^h$ is the attention distribution produced by the $h$-th attention head. Finally, the output states are concatenated to produce the final state.

\section{Approach}

Multi-head attention allows the model to jointly attend to information from {\em different} representation subspaces at {\em different} positions. 
To further guarantee the diversity,
we enlarge the distances among multiple attention heads with disagreement regularization (Section~\ref{sec:framework}). 
Specifically, we propose three types of disagreement regularization to encourage each head vector $O^i$ to be different from other heads (Section~\ref{sec-regularization}).


\subsection{Framework}
\label{sec:framework}
 In this work, we take the machine translation tasks as application. Given a source sentence ${\bf x}$ and its translation ${\bf y}$, a neural machine translation model is trained to maximize the conditional translation probability over a parallel training corpus.
 
 We introduce an auxiliary regularization term in order to encourage the diversity among multiple attention heads.
Formally, the training objective is revised as:
\begin{equation}
J(\theta) = \argmax_{\theta} \big\{\underbrace{L({\bf y}|{\bf x}; \theta)}_\text{\normalsize \em likelihood} + \lambda *\underbrace{D({\bf a} | {\bf x}, {\bf y}; \theta)}_\text{\normalsize \em disagreement} \big\}, \label{eq:framework} \nonumber
\end{equation}
where ${\bf a}$ is the referred attention matrices,  $\lambda$ is a hyper-parameter and is empirically set to 1.0 in this paper. The auxiliary regularization term $D(\cdot)$ guides the related attention component to capture different features from the corresponding projected subspaces.

Note that the introduced regularization term works like $L1$ and $L2$ terms, which do not introduce any new parameters and only influence the training of the standard model parameters.

\subsection{Disagreement Regularization}
\label{sec-regularization}

Three types of regularization term, which are applied to three parts of the original multi-head attention, are introduced in this section.

\paragraph{\bf Disagreement on Subspaces (Sub.)} This disagreement is designed to maximize the cosine distance between the projected values. Specifically, we first calculate the cosine similarity $\cos(\cdot)$ between the vector pair $V^i$ and $V^j$ in different value subspaces, through the dot product of the normalized vectors\footnote{We did not employ the Euler Distance between vectors since we do not care the absolute value in each vector.}, which measures the cosine of the angle between $V^i$ and $V^j$. Thus, the cosine distance is defined as negative similarity, i.e, $-\cos(\cdot)$. Our training objective is to enlarge the average cosine distance among all head pairs. The regularization term is formally  expressed as:
\begin{equation}\label{eq:sub}
D_{subpace} = - \frac{1}{H^2} \sum_{i=1}^{H}\sum_{j=1}^{H} \frac{V^i \cdot V^j}{\|V^i\|\|V^j\|}.
\end{equation}

\paragraph{\bf Disagreement on Attended Positions (Pos.)} Another strategy is to disperse the attended positions predicted by multiple heads.  Inspired by the agreement regularization~\cite{Liang:2006:NAACL,Cheng:2016:IJCAI} which encourages multiple alignments to be similar, in this work, we deploy a variant of the original term by introducing an alignment disagreement regularization. 
Formally, we employ the sum of element-wise multiplication of corresponding matrix cells\footnote{We also used the squared element-wise subtraction of two matrices in our preliminary experiments, and found it underperforms its multiplication counterpart, which is consistent with the results in ~\cite{Cheng:2016:IJCAI}.}, to measure the similarity between two matrices $A^i$ and $A^j$ of two heads:
\begin{equation}\label{eq:pos}
	D_{position} = - \frac{1}{H^2} \sum_{i=1}^{H}\sum_{j=1}^{H} |A^i \odot A^j|.
\end{equation}

\paragraph{\bf Disagreement on Outputs (Out.)} This disagreement directly applies regularization on the outputs of each attention head, by maximizing the difference among them. 
Similar to the \emph{subspace} strategy, we employ negative cosine similarity to measure the distance:
\begin{equation}\label{eq:out}
D_{output} = - \frac{1}{H^2} \sum_{i=1}^{H}\sum_{j=1}^{H} \frac{O^i \cdot O^j}{\|O^i\|\|O^j\|}.
\end{equation}

\section{Related Work}
The regularization on attended positions is inspired by agreement learning in prior works, which encourages alignments or hidden variables of multiple models to be similar.
\newcite{Liang:2006:NAACL} first assigned agreement terms for jointly training word alignment in phrase-based statistic machine translation \cite{koehn2003statistical}. The idea was further extended into other natural language processing tasks such as grammar induction \cite{liang2008agreement}. \newcite{Levinboim:2015:NAACL} extended the agreement for general bidirectional sequence alignment models with model inevitability regularization. \newcite{Cheng:2016:IJCAI} further explored the agreement on modeling the source-target and target-source alignments in neural machine translation model. In contrast to the mentioned approaches which assigned agreement terms into loss function, we deploy an alignment disagreement regularization by maximizing the distance among multiple attention heads.

As standard multi-head attention model lacks effective control on the influence of different attention heads, \newcite{ahmed2017weighted} used a weighted mechanism to combine them rather than simple concatenation. 
As an alternative approach to multi-head attention, \newcite{Shen:2018:AAAI} extended the single relevance score to multi-dimensional attention weights, demonstrating the effectiveness of modeling multiple features for attention networks.
Our approach is complementary to theirs: our model encourages the diversity among multiple heads, while theirs enhance the power of each head.

\section{Experiments}

\subsection{Setup}

To compare with the results reported by previous work~\cite{pmlr-v70-gehring17a,Vaswani:2017:NIPS,hassan2018achieving}, we conduct experiments on both WMT2017 Chinese$\Rightarrow$English (Zh$\Rightarrow$En) and WMT2014 English$\Rightarrow$German (En$\Rightarrow$De) translation tasks.
The Zh$\Rightarrow$En corpus consists of 20M sentence pairs, and the En$\Rightarrow$De corpus consists of 4M sentence pairs. We follow previous work to select the validation and test sets.
Byte-pair encoding (BPE) is employed to alleviate the Out-of-Vocabulary problem \cite{sennrich2016neural} with 32K merge operations for both language pairs.
We use the case-sensitive 4-gram NIST BLEU score~\cite{papineni2002bleu} as evaluation metric, and \emph{sign-test}~\cite{Collins:2005} for statistical significance test.

We evaluate the proposed approaches on the advanced \textsc{Transformer} model~\cite{Vaswani:2017:NIPS}, and implement on top of an open-source toolkit -- THUMT~\cite{zhang2017thumt}. We follow~\newcite{Vaswani:2017:NIPS} to set the configurations and have reproduced their reported results on the En$\Rightarrow$De task. 
All the evaluations are conducted on the test sets.
We have tested both \emph{Base} and \emph{Big} models, which differ at hidden size (512 vs. 1024) and number of attention heads (8 vs. 16). We study model variations with \emph{Base} model on the Zh$\Rightarrow$En task (Section \ref{sec:terms} and \ref{sec:component}), and evaluate overall performance with \emph{Big} model on both Zh$\Rightarrow$En and En$\Rightarrow$De tasks (Section \ref{sec:comparision}).

\begin{table}[t]
  \centering
  \begin{tabular}{c|c|c|c||c||c}
    \multirow{2}{*}{\#}  & \multicolumn{3}{c||}{\bf Regularization}  &  \multirow{2}{*}{\bf Speed}   &   \multirow{2}{*}{\bf BLEU} \\  
    \cline{2-4}
                        & \em Sub. &   \em Pos.   &   \em Out.  &   \\
    \hline
    1   &   \texttimes   &   \texttimes   &   \texttimes  & 1.21  &  24.13\\
    \hline
    \hline
    2   &   \checkmark  &   \texttimes  &   \texttimes  & 1.15  &   24.64\\
    3   &   \texttimes  &   \checkmark  &   \texttimes  &  1.14 &   24.42\\
    4   &   \texttimes  &   \texttimes  &   \checkmark  & 1.15  &   \bf 24.78\\
    \hline
    5   &   \checkmark  &   \texttimes  &   \checkmark  & 1.12 &   24.73 \\
    6   &   \checkmark  &   \checkmark  &   \texttimes  &1.11&   24.38 \\
    7   &   \checkmark  &   \checkmark  &   \checkmark  &1.05 &   24.60 \\
  \end{tabular}
  \caption{Effect of regularization terms, which are applied to the encoder self-attention only. ``Speed'' denotes the training speed (steps/second).}
  \label{tab:term}
\end{table}

\begin{table*}[t]
  \setcounter{table}{2}
  \centering
  \begin{tabular}{l|l||rl|rl}
    \multirow{2}{*}{\bf System}  &   \multirow{2}{*}{\bf Architecture}  & \multicolumn{2}{c}{\bf Zh$\Rightarrow$En}  &  \multicolumn{2}{|c}{\bf En$\Rightarrow$De}\\
    \cline{3-6}
        &   &   Speed    &   BLEU    &   Speed    &   BLEU\\
    \hline \hline
    \multicolumn{6}{c}{{\em Existing NMT systems}} \\
    \hline
    \cite{wu2016google} &   \textsc{GNMT}             &  n/a &  n/a   &   n/a &   26.30\\ 
    \cite{pmlr-v70-gehring17a}  &   \textsc{ConvS2S}  &  n/a &  n/a   &   n/a &   26.36\\
    \hline
    \multirow{2}{*}{\cite{Vaswani:2017:NIPS}} &   \textsc{Transformer-Base}    &    n/a & n/a &  n/a &   27.3\\ 
    &  \textsc{Transformer-Big}               &  n/a  &  n/a  &  n/a &  28.4\\ 
    \hdashline
    \cite{hassan2018achieving}  &   \textsc{Transformer-Big}  &  n/a  &  24.2  &  n/a  & n/a\\
    \hline\hline
    \multicolumn{6}{c}{{\em Our NMT systems}}   \\ \hline
    \multirow{4}{*}{\em this work}  &   \textsc{Transformer-Base}   &1.21 & 24.13    &  1.28  &  27.64   \\
    &   ~~~ + Disagreement  &  1.06 &   24.85$^\Uparrow$   & 1.10 &   28.51$^\Uparrow$  \\
    \cline{2-6}
    &   \textsc{Transformer-Big} & 0.58 & 24.56    & 0.61&  28.58   \\
    &   ~~~ + Disagreement       & 0.47 &  25.08$^\Uparrow$   & 0.51  &   29.28$^\Uparrow$ \\  
  \end{tabular}
  \caption{Comparing with existing NMT systems on WMT17 Chinese$\Rightarrow$English and WMT14 English$\Rightarrow$German translation tasks. ``$\Uparrow$'' indicates that the model is significantly better than its baseline counterpart ($p < 0.01$).}
  \label{tab:exist}
\end{table*}

\subsection{Effect of Regularization Terms}\label{sec:terms}
In this section, we evaluate the impact of different regularization terms on the  Zh$\Rightarrow$En task using \textsc{Transformer-Base}. For simplicity and efficiency, here we only apply regularizations on the encoder side. As shown in Table~\ref{tab:term}, 
all the models with the proposed disagreement regularizations (Rows 2-4) consistently outperform the vanilla \textsc{Transformer} (Row 1). Among them, the \emph{Output} term performs best which is +0.65 BLEU score better than the baseline model, the \emph{Position} term is less effective than the other two.
In terms of training speed, we do not observe obvious decrease, which in turn demonstrates the advantage of our disagreement regularizations.

However, the combinations of different disagreement regularizations fail to further improve translation performance (Rows 5-7). One possible reason is that different regularization terms have overlapped guidance, and thus combining them does not introduce too much new information while makes training more difficult. 




\begin{table}[t]
  \setcounter{table}{1}
  \centering
  \begin{tabular}{c|c|c||c||c}
    \multicolumn{3}{c||}{\bf Applying to}  &  \multirow{2}{*}{\bf Speed}   &   \multirow{2}{*}{\bf BLEU}  \\  
    \cline{1-3}
    \em Enc &   \em E-D   &   \em Dec  &   &\\
    \hline
    \texttimes  &   \texttimes   &  \texttimes  &  1.21 &   24.13\\
    \hline
    \checkmark  &   \texttimes   &  \texttimes  & 1.15  &   24.78\\
    \checkmark  &   \checkmark   &  \texttimes  & 1.10  &   24.67\\
    \checkmark  &   \texttimes   &  \checkmark  & 1.11 &   24.69\\
    \checkmark  &   \checkmark   &  \checkmark  & 1.06  &  \textbf{24.85}\\
  \end{tabular}
  \caption{Effect of regularization on different attention networks, i.e., encoder self-attention (``{\em Enc}''), encoder-decoder attention (``{\em E-D}''), and decoder self-attention (``{\em Dec}'').} 
  \label{tab:component}
\end{table}

\subsection{Effect on Different Attention Networks}
\label{sec:component}

The \textsc{Transformer} consists of three attention networks, including encoder self-attention, decoder self-attention, and encoder-decoder attention. 
In this experiment, we investigate how each attention network benefits from the disagreement regularization. 
As seen from Table~\ref{tab:component}, all models consistently improve upon the baseline model.
When applying disagreement regularization to all three attention networks, we achieve the best performance, which is +0.72 BLEU score better than the baseline model. 
The training speed decreases by 12\%, which is acceptable considering the performance improvement.

\subsection{Main Results}
\label{sec:comparision}

Finally, we validate the proposed disagreement regularization on both WMT17 Chinese-to-English and WMT14 English-to-German translation tasks. 
Specifically, we adopt the \emph{Output} disagreement regularization, which is applied to all three attention networks. 
The results are concluded in Table \ref{tab:exist}. We can see that our implementation of \textsc{Transformer} outperforms all existing NMT systems, and matches the results of \textsc{Transformer} reported in previous works.
Incorporating disagreement regularization consistently improves translation performance for both base and big \textsc{Transformer} models across language pairs, demonstrating the effectiveness of the proposed approach.
It is encouraging to see that \textsc{Transformer-Base} with disagreement regularization achieves comparable performance with \textsc{Transformer-Big}, while the training speed is nearly twice faster.


\begin{table}[t]
  \centering
  \setcounter{table}{3}
    \begin{tabular}{c|c|c|c}
    \multirow{2}{*}{\bf Regularization on}  & \multicolumn{3}{c}{\bf Disagreement on}\\  
    \cline{2-4}
                        & \em Sub. &   \em Pos.   &   \em Out.    \\
    \hline
    n/a   &   0.882   &   0.007   &   0.881 \\
    \hline
    \hline
    Subspace   &   \bf 0.999  &   0.006  &   0.935  \\
    Position   &   0.882  &   \bf 0.219  &   0.882  \\
    Output   &   0.989  &   0.006  &   \bf 0.997  \\
  \end{tabular}
  \caption{Effect of different regularization terms on the three disagreement measurements. ``n/a'' denotes the baseline model without any regularization term. Larger value denotes more disagreement (at most 1.0).}
  \label{tab:analysis}
\end{table}

\begin{table}[t]
  \centering
  \scalebox{0.84}{
  \begin{tabular}{c||c|c|c|c|c|c}
    \multirow{2}{*}{\bf Reg.}  & \multicolumn{6}{c}{\bf Layer}\\  
    \cline{2-7}
                        & $1$ & $2$ &   $3$ &   $4$ &   $5$ &   $6$\\
    \hline
    n/a   &   0.040   &   0.009   &   0.002 &   0.003   &   0.008   &   0.006\\
    \hline
    \hline
    Sub.   &   0.039    &   0.009   &   0.001   &   0.003   &   0.006   &   0.005\\
    Pos.   &   0.217    &   0.167   &   0.219   &   0.242   &   0.233   &   0.249\\
    Out.   &   0.048    &   0.009   &   0.002   &   0.003   &   0.008   &   0.006\\
  \end{tabular}
  }
  \caption{Disagreement on attended positions with respect to the levels of the encoder layers.}
  \label{tab:pos}
\end{table}

\subsection{Quantitative Analysis of Regularization}

In this section, we empirically investigate how the regularization terms affect the multi-head attention. To this end, we compare the disagreement scores on subspaces (``Sub.''), attended positions (``Pos.''), and outputs (``Out.''). Since the scores are negative values, we list $\exp(D)$ for readability, which has a maximum value of 1.0. 
Table~\ref{tab:analysis} lists the results of encoder-side multi-head attention on the Zh$\Rightarrow$En validation set.
As seen, the disagreement score on the individual component indeed increases with the corresponding regularization term. For example, the disagreement of outputs increases to almost 1.0 by using the \emph{Output} regularization, which means that the output vectors are almost perpendicular to each other as we measure the cosine distance as the disagreement.

One interesting finding is that attending to different positions may not be the essential strength of multi-head attention on the translation task. As seen, the disagreement score on the attended positions for the standard multi-head attention is only 0.007, which indicates that almost all the heads attend to the same positions. 
Table~\ref{tab:pos} shows the disagreement scores on attended positions across encoder layers. Except for the $1^{st}$ layer that attends to the input word embeddings, the disagreement scores on other layers (i.e. ranging from the $2^{nd}$ to $6^{th}$ layer) are very low, which confirms out above hypothesis.

Concerning the regularization terms, except that on position, the other two regularization terms (i.e. ``Sub.'' and ``Out.'') do not increase the disagreement score on the attended positions. This can explain why positional regularization term does not work well with the other two terms, as shown in Table~\ref{tab:term}. This is also consistent with the finding in~\cite{tu2016modeling}, which indicates that neural networks can model linguistic information in their own way. In contrast to attended positions, it seems that the multi-head attention prefer to encoding the differences among multiple heads in the learned representations.



\section{Conclusion}
In this work, we propose several disagreement regularizations to augment the multi-head attention model, which encourage the diversity among attention heads so that different head can learn distinct features. Experimental results across language pairs validate the effectiveness of the proposed approaches.

The models also suggest a wide range of potential advantages and extensions, from being able to improve the performance of multi-head attention in other tasks such as reading comprehension and language inference, to being able to combine with other techniques~\cite{shaw2018self,Shen:2018:AAAI,Dou:2018:EMNLP,Yang:2018:EMNLP} to further improve performance.


\section*{Acknowledgments}
The work was supported by the National Natural Science Foundation of China (Project No. 61332010 and No. 61472338), the Research Grants Council of the Hong Kong Special Administrative Region, China (No. CUHK 14234416 of the General Research Fund), and Microsoft Research Asia (2018 Microsoft Research Asia Collaborative Research Award).
We thank the anonymous reviewers for their insightful comments.

\balance
\bibliographystyle{acl_natbib}
\bibliography{ref} 
\end{document}